\documentclass[sigconf]{acmart}

\usepackage{amsmath}
\usepackage{booktabs}
\usepackage{makecell}
\usepackage{multirow}
\usepackage{subfig} 
\usepackage{mathrsfs}
\usepackage{array}  
\usepackage{subcaption} 
\usepackage{marvosym}

\AtBeginDocument{%
  }

\setcopyright{acmlicensed}
\copyrightyear{2026}
\acmYear{2026}
\setcopyright{cc}
\setcctype{by}
\acmConference[WWW '26]{Proceedings of the ACM Web Conference 2026}{April 13--17, 2026}{Dubai, United Arab Emirates}
\acmBooktitle{Proceedings of the ACM Web Conference 2026 (WWW '26), April 13--17, 2026, Dubai, United Arab Emirates}
\acmPrice{}
\acmDOI{10.1145/3774904.3793059}
\acmISBN{979-8-4007-2307-0/2026/04}

\begin{document}

\title{WED-Net: A Weather-Effect Disentanglement Network with Causal Augmentation for Urban Flow Prediction}

\author{Qian Hong}    
\affiliation{%
  \institution{Gaoling School of Artificial Intelligence, Renmin University of China}
  \city{Beijing}
  \country{China}}
\email{qianhong99@ruc.edu.cn}
 
\author{Siyuan Chang}
\affiliation{%
   \institution{School of Statistics, Renmin University of China}
  \city{Beijing}
  \country{China}}
\email{csy24@ruc.edu.cn}

\author{Xiao Zhou}
\authornote{Corresponding authors}
\authornote{Also with Beijing Key Laboratory of Research on Large Models and Intelligent Governance}
\authornote{Also with Engineering Research Center of Next-Generation Intelligent Search and Recommendation, MOE}

\affiliation{%
  \institution{Gaoling School of Artificial Intelligence, Renmin University of China}
  \city{Beijing}
  \country{China}}
\email{xiaozhou@ruc.edu.cn}



\begin{abstract}
  Urban spatio-temporal prediction under extreme conditions (e.g., heavy rain) is challenging due to event rarity and dynamics.  Existing data-driven approaches that incorporate weather as auxiliary input often rely on coarse-grained descriptors and lack dedicated mechanisms to capture fine-grained spatio-temporal effects. Although recent methods adopt causal techniques to improve out-of-distribution generalization, they typically overlook temporal dynamics or depend on fixed confounder stratification. To address these limitations, we propose \textbf{WED-Net} (Weather-Effect Disentanglement Network), a dual-branch Transformer architecture that separates intrinsic and weather-induced traffic patterns via self- and cross-attention, enhanced with memory banks and fused through adaptive gating. To further promote disentanglement, we introduce a discriminator that explicitly distinguishes weather conditions. Additionally, we design a causal data augmentation strategy that perturbs non-causal parts while preserving causal structures, enabling improved generalization under rare scenarios. Experiments on taxi-flow datasets from three cities demonstrate that WED-Net delivers robust performance under extreme weather conditions, highlighting its potential to support safer mobility, highlighting its potential to support safer mobility, disaster preparedness, and urban resilience in real-world settings. The code is publicly available at \href{https://github.com/HQ-LV/WED-Net}{https://github.com/HQ-LV/WED-Net}.
  
\end{abstract}

\ccsdesc[500]{Computing methodologies~Machine learning}
\ccsdesc[500]{Information systems~Spatial-temporal systems}
\ccsdesc[500]{Information systems~Data mining}

\keywords{Traffic forecasting, Spatio-temporal models, Disaster response}


\maketitle

\begin{figure}[ht]
\centering
\includegraphics[width=0.99\columnwidth]{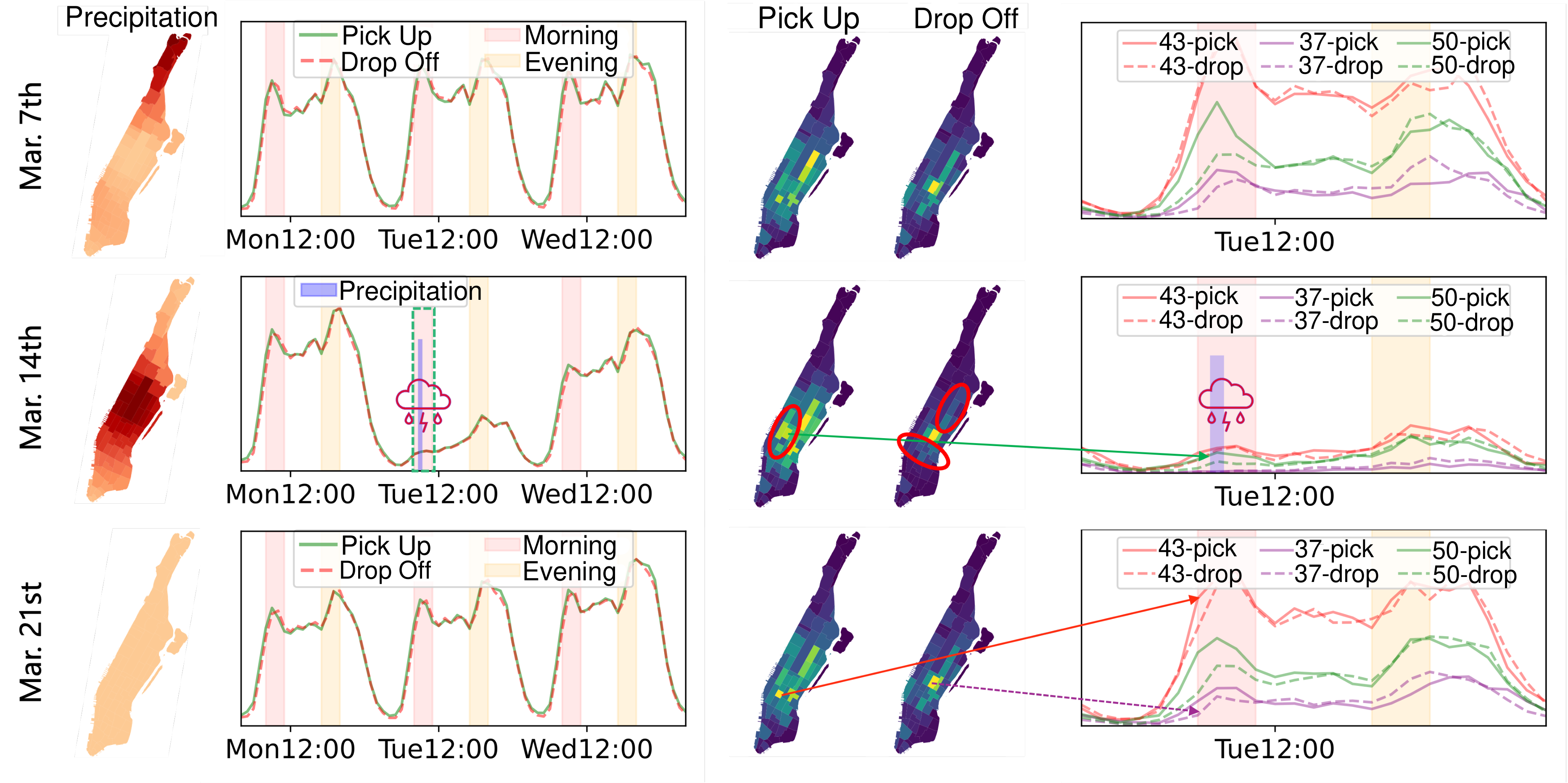}  
\caption{Comparison of spatio-temporal patterns of NYC taxi flow on Mar. 7th, Mar. 14th (rainstorm), and Mar. 21st. }
\label{intro-case}
\end{figure}

\section{Introduction}  
Urban flow prediction is a cornerstone of smart city development, as it enables the optimization of traffic management systems and long-term urban planning, supports timely emergency response and disaster management, and provides quantitative insights into underlying social and mobility dynamics. Classical deep learning models such as STGCN~\cite{10.5555/3304222.3304273}, DCRNN~\cite{li2018diffusion}, Graph WaveNet~\cite{10.5555/3367243.3367303}, and  STAEformer~\cite{10.1145/3583780.3615160} leverage spatio-temporal graph neural networks and Transformer-based architectures to capture complex spatial correlations and temporal dependencies. 
By learning expressive representations of evolving traffic states, these methods have advanced applications such as route planning, travel time estimation, congestion mitigation, and reliable intelligent transportation services in dynamic urban environments.

Nevertheless, existing spatio-temporal models either completely overlook or only coarsely encode external factors such as extreme weather, large public events, and other disruptive disasters, even though these exogenous influences can drastically alter normal traffic patterns and substantially increase temporal and spatial volatility. 
When these factors are ignored or treated as noise, models become biased toward typical conditions, leading to inaccurate predictions and delayed responses when reliable forecasts are most needed.
 
To illustrate the substantial spatio-temporal variation of human mobility under different weather conditions, we examine taxi flow in New York City (NYC) across three Tuesdays (March 7th, 14th, and 21st), as shown in Fig.~\ref{intro-case}. Light red and yellow shaded intervals indicate the morning (8:00 to 11:00 AM) and evening (5:00 to 8:00 PM) peak hours, respectively, thereby highlighting the two main commuting periods. The line plots in the left column show that the rainstorm on March 14th leads to a clear and sustained drop in total flow throughout the day, which differs noticeably from the two rain-free Tuesdays that display relatively similar and stable trajectories of overall demand. The middle column presents the spatial distributions of taxi flow during the morning peak across the three Tuesdays, and evident changes emerge in the circled regions when March 14th experienced heavy rainfall, indicating that not only the overall volume but also the spatial allocation of trips is affected. The parcel-level trends (IDs 37, 43, and 50) shown in the right-column line plots indicate that, under normal conditions, the three parcels exhibit similar temporal patterns with consistent relative magnitudes over time. However, the rainstorm disrupts this pattern, causing parcel 43’s flow to drop sharply to approximately the same level as that of parcel 50.

The aforementioned classical urban flow prediction models typically overlook the spatially heterogeneous nature of weather and lack dedicated components to capture its effects in a principled manner. In many existing approaches, external weather variables are treated analogously to temporal covariates such as daily or weekly periodicity and holidays: they are incorporated as auxiliary inputs that are shared across all time steps or spatial regions and are simply concatenated with the traffic features. For example, MemeSTN and EAST-Net~\cite{jiang2023learning,wang2022event} construct memory banks of traffic patterns that are stratified by coarse weather-type labels (e.g., normal vs. disaster days), while MT-C2G~\cite{10.1145/3557915.3561029} addresses data imbalance by oversampling traffic samples tagged as extreme-weather cases. However, these approaches do not directly model the fine-grained weather observations and instead treat external events as coarse, qualitative labels attached to traffic data, rather than as continuously varying, spatially heterogeneous factors. As a result, their ability to capture detailed, localized impacts of weather on traffic flow remains limited. In practice, different precipitation levels and their spatially uneven distribution can influence travel behavior in markedly different ways: a sudden moderate rainfall event may temporarily increase taxi demand in specific areas as travelers switch from walking or cycling to motorized modes, whereas a well-forecasted severe storm can suppress demand throughout the day or shift peak travel period earlier or later across the city as individuals adjust their schedules, with some neighborhoods being more strongly affected than others.  These nuanced, fine-grained weather effects are often overlooked, leaving models unable to capture their complex influence on traffic patterns.

Moreover, the above purely data-driven approaches often learn spurious correlations, particularly under rare external events such as heavy rainfall. As illustrated in Fig.~\ref{intro-case}, extreme weather can induce substantial shifts in both the spatial distribution and temporal evolution of traffic patterns. Because such events occur infrequently, they are severely underrepresented in the training data, which in turn leads to systematic prediction failures when they do arise~\cite{wang2016patterns,10.1145/2623330.2623628}. These limitations motivate our focus on robust out-of-distribution (OOD) prediction under dynamic external conditions.

In graph learning and computer vision, OOD generalization has been approached through methods~\cite{li2022let,zhuang2023learning,liu2023flood,mao2022causal} grounded in the principle of \textit{Invariance of Causality}~\cite{buhlmann2020invariance}, which posits that causal relationships remain stable across different environments. Recent studies in spatio-temporal learning have begun to adopt causal tools to address distribution shifts in a similar spirit~\cite{wang2024nuwadynamics,duan2024causal,wang2024stone,ji2023self,li2025spatio}. These methods either perform data augmentation by identifying causal and non-causal regions and perturbing the latter in the input or latent space, or apply causal adjustment techniques to reduce the influence of environmental confounders. Moreover, temporal causality is often overlooked. These methods may distort underlying causal structures with poorly designed perturbations, as well as rely on fixed confounders that are ill-suited to dynamic urban traffic under changing weather.

Motivated by the above insights, we propose \textbf{WED-Net} (Weather-Effect Disentanglement Network), a forecasting framework that explicitly separates intrinsic traffic patterns from weather-induced dynamics. WED-Net adopts a dual-branch Transformer architecture: \textit{I-STEnc} employs spatio-temporal self-attention to capture intrinsic traffic dependencies, while \textit{W-STEnc} leverages cross-attention to model fine-grained weather effects on traffic states. Both branches are equipped with memory banks that retrieve representative historical patterns and are integrated through an adaptive gating mechanism that assigns data-dependent weights to each branch. To further promote disentanglement, a \textit{Weather Discriminator} is trained in an adversarial manner to encourage weather-invariant representations in the intrinsic branch.
To improve robustness under rare conditions, we further propose a \textit{causality-driven augmentation} strategy that perturbs non-causal parts while preserving causal ones. In contrast to prior causal augmentation methods, our approach jointly considers temporal and spatial causality and is grounded in real-world meteorological semantics, which helps maintain meaningful causal structures and enhances both interpretability and generalization under distribution shifts. Extensive experiments on three real-world city datasets show that WED-Net improves generalization under extreme weather conditions and strengthens the capacity of traffic forecasting systems to support disaster-response applications.

Our contributions are summarized as follows:
\begin{itemize} 
    \item We model fine-grained weather effects on traffic with a weather-aware spatio-temporal network that disentangles intrinsic and weather-induced patterns for robust prediction under dynamic conditions.
    \item We introduce a causality-driven augmentation method that perturbs non-causal parts to improve generalization under rare extreme weather scenarios.
    \item We validate our approach on real-world datasets from three cities, demonstrating superior performance in dynamic traffic prediction and disaster response scenarios.
\end{itemize}

\section{Related Work}

\noindent\textbf{Traffic Forecasting.}
Capturing complex spatio-temporal dependencies is a key challenge in traffic prediction.  Spatio-temporal graph models like STGCN~\cite{10.5555/3304222.3304273} and DCRNN~\cite{li2018diffusion} combine TCN and GCN modules to model spatio-temporal features. ASTGCN~\cite{guo2019attention} proposes using daily and weekly periodicity to enhance temporal modeling. To improve static modeling of predefined graph structures, models like Graph WaveNet~\cite{10.5555/3367243.3367303}, MTGNN~\cite{10.1145/3394486.3403118}, GTS~\cite{shang2021discrete}, and AGCRN~\cite{bai2020adaptive} learn dynamic and adaptive network structures from data. More Recently, Transformers have also become primary architecture for traffic  forecasting. Models like STAEformer~\cite{10.1145/3583780.3615160} can capture adaptive spatial structures and dynamic temporal dependencies. Meanwhile,  Autoformer~\cite{wu2021autoformer}, and PDFormer~\cite{10.1609/aaai.v37i4.25556} focus on long sequence event prediction.

\noindent\textbf{Dynamic Urban Flow Prediction.}
However, existing researches have not sufficiently analyzed the impact of external events (e.g., urban large-scale events, traffic accidents, extreme weather, and natural disasters) on human mobility. Quantitative empirical studies have shown that human mobility patterns under disruptions from natural disasters, such as hurricanes and earthquakes, are related to steady-state mobility patterns~\cite{wang2016patterns,10.1145/2623330.2623628}.   
Based on this discovery, recent models such as DIGC-Net~\cite{10.1145/3340531.3411873}, MT-C2G~\cite{10.1145/3557915.3561029}, and EAST-Net~\cite{wang2022event} leverage deep models to capture the effects of external factors like events and weather on human mobility and urban traffic. However, such external factor data is sparse, and factors like weather are often treated as globally shared external factors, concatenated with traffic data before being input to the network. In reality, the impact of weather also exhibits spatio-temporal heterogeneity, which is often overlooked in existing studies.

\noindent\textbf{Causal Solutions for OOD Generalization.}
The scarcity of traffic data under extreme weather conditions poses a typical out-of-distribution (OOD) challenge, where spatio-temporal dependencies learned from normal scenarios fail to generalize. In graph learning and computer vision, OOD generalization is tackled by methods that leverage the invariance of causality~\cite{buhlmann2020invariance}, which assumes that causal relationships remain stable across different environments. Models like RGCL~\cite{li2022let}, iMoLD~\cite{zhuang2023learning}, and FLOOD~\cite{liu2023flood} learn causal representations of graph by generating perturbed but consistent views. In image recognition, Mao et al.~\cite{mao2022causal} separate causal content from spurious background and apply causal transportability to improve robustness under distribution shifts. Recently, spatio-temporal studies like NuwaDynamics~\cite{wang2024nuwadynamics}, CaPaint~\cite{duan2024causal}, STONE~\cite{wang2024stone}, and STEVE~\cite{ji2023self} have explored causal tools for OOD robustness. However, these methods often overlook temporal causality, rely on simplistic spatio-temporal mechanisms that may distort causal structures, and depend on fixed confounders that struggle to adapt to changing environments.

To mitigate these shortcomings, we propose a framework that disentangles intrinsic spatio-temporal traffic dependencies from weather-induced patterns with fine-grained weather modeling, and incorporates spatio-temporal causal augmentation to better capture traffic dynamics under extreme weather.

\begin{figure*}[!ht]
\centering
\includegraphics[width=0.94\textwidth]{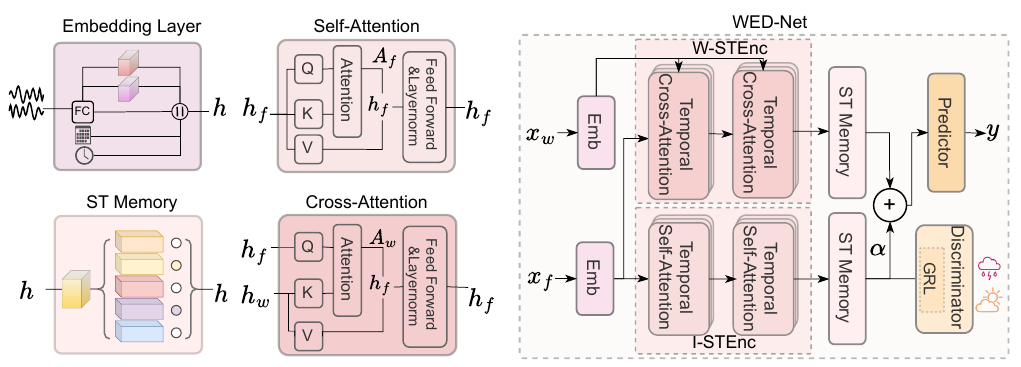}  
\caption{The architecture of WED-Net. Embedded traffic and weather inputs are processed by \textit{I-STEnc} and \textit{W-STEnc} to disentangle intrinsic dynamics and weather-induced effects via attention. \textit{ST Memory} enhances each branch with historical patterns, a \textit{Weather Discriminator} enforces weather invariance, and \textit{Adaptive Fusion} combines both branches for prediction.}
\label{framework}
\end{figure*}

\section{Preliminaries}

This section establishes the key concepts, notation, and the core research problem of weather-aware spatio-temporal (ST) prediction.

\noindent \textbf{{Urban Region Graph.}} 
Given a city, we define an \emph{ Urban Region Graph} $\mathcal{G} (\mathcal{V},\mathcal{E},\mathcal{A})$, 
where nodes $\mathcal{V}$ represent land parcels, edges $\mathcal{E}$ connect adjacent parcels, and the adjacency matrix $\mathcal{A}$ represents the geographical distance between parcels.

\noindent \textbf{{Urban Flow.}}   
Given graph $\mathcal{G}$ and a time window $L$, $X \in \mathbb{R}^{L \times N \times d_f}$ denotes flow data, where $d_f$ is the number of flow features and $N = |\mathcal{V}|$ is the number of spatial units.

\noindent \textbf{{Meteorological Data.}} 
Given $\mathcal{G}$ and $L$, the meteorological data is denoted by $M \in \mathbb{R}^{L \times N \times {d_m}}$, with ${d_m}$ is the number of meteorological attributes, including hourly rainfall, temperature, wind velocity.

\noindent \textbf{{Weather-Aware Spatio-Temporal Prediction.}}
Given urban flow  $X_{t-T+1:t}$ and corresponding meteorological data $M_{t-T+1:t}$ in the previous $T$ time frames, weather-aware spatio-temporal forecasting aims to infer the urban flow in the future  $T'$ frames by training a model $\mathcal{F}(\cdot)$ with parameters $\theta$, which can be formulated as:
\begin{align}
\left[X_{t-T+1:t}, M_{t-T+1:t}  \right]  \overset{\mathcal{F}(\cdot)}{\underset{\theta}{\longrightarrow}  }
 X_{t+1:t+T'} .
\end{align}

\section{Methodology}
\subsection{Weather-Effect Disentanglement Network}  
As shown in Fig.~\ref{framework}, WED-Net embeds raw traffic and weather inputs, then applies a two-branch architecture with \textit{I-STEnc} and \textit{W-STEnc} that leverage self- and cross-attention to disentangle intrinsic traffic dynamics from weather-induced variations. Each branch is enhanced with \textit{ST Memory} for historical pattern retrieval, while a \textit{Weather Discriminator} enforces weather-invariant intrinsic features, and \textit{Adaptive Fusion} combines both branches for accurate, context-aware forecasting.

\subsubsection{Embedding Layer. }
For raw traffic or weather data $F \in \mathbb{R}^{T \times N \times  d }  $, we first use a fully connected layer to obtain feature embeddings $E_h = FC(F)$. Then we project them into adaptive embeddings $E_{a_t} $ and $E_{a_s}$ using time-adaptive and spatial-adaptive weight matrices, respectively. To capture daily and weekly periodicity, we store the time-of-day and day-of-week features using one-hot embeddings, and use the time indices of the raw data to extract periodic representations $E_p$. By concatenating these embeddings, we obtain the initial hidden representation $h \in \mathbb{R}^{T \times N \times  d_{h}} $.

\subsubsection{Intrinsic and Weather-induced ST Dependency Disentanglement.}
Urban traffic dynamics are governed by heterogeneous factors, primarily stemming from two sources: intrinsic spatio-temporal dependencies, such as daily commuting patterns and urban functional layouts; and external perturbations, particularly weather events like rain or snow, which can induce abrupt and localized disruptions. Existing models often entangle these factors within a unified representation, making it difficult to generalize under extreme or shifting weather conditions. To address this, we propose a structural disentanglement of these two dependencies.
Specifically, we design a dual-branch module:
\begin{itemize}
    \item The Intrinsic Spatio-Temporal Dependency Encoder (I-STEnc) employs self-attention transformer to model the internal spatio-temporal correlations within traffic data.
    \item The Weather-Induced Spatio-Temporal Dependency Encoder (W-STEnc) leverages cross-attention, treating traffic features as queries and weather features as keys and values, to model weather’s modulation of traffic patterns.
\end{itemize}
This explicit disentanglement enables distinct representations for intrinsic and environmental dependencies, improving sensitivity to weather shifts and generalization under diverse conditions.

In \textit{I-STEnc}, temporal and spatial self-attention blocks model intrinsic urban flow dependencies. Given hidden embeddings $h_f$, we obtain queries, keys, and values along the temporal axis:
\begin{equation} 
 Q_{f}^{t} ={h_f} W_{Q_{f}}^{t}, 
 K_{f}^{t} ={h_f} W_{K_{f}}^{t}, 
 V_{f}^{t} ={h_f} W_{V_{f}}^{t}, 
\end{equation}
then compute the temporal self-attention scores: 
\begin{equation}
A_{f}^{t } = \text{Softmax} \left(\frac{Q_{f}^{ t } K_{f}^{t\top}}{\sqrt{d_h}}\right),
\end{equation} 
where temporal attention $A^{t}_{f} \in \mathbb{R}^{N \times T \times T}$ is the dependencies between different time steps. Finally, we use the temporal attention scores to weight the hidden inputs, obtaining the output of the temporal self-attention  $h^t_f= A_{f}^{t} V_{f}^{t} $. We also apply feed forward, layer normalization, residual connection and multi-head mechanism. 
Similarly, we obtain the spatial transformer block representation $h^s_f= A_{f}^{s} V_{f}^{s} $, where the spatial attention $A_{f}^{s} \in \mathbb{R}^{T \times N \times N}$ represents the dependencies between different spatial parcels.

We denote above spatio-temporal self-attention block as:
\begin{equation}
    h_f^{\text{intr}} = \text{SelfAttn}({h_f}).
\end{equation}

In \textit{W-STEnc}, we use cross-attention mechanism to capture fine-grained spatio-temporal impact of weather on urban mobility: 
\begin{equation}
    Q_{w}^a=h_f W_{Q_{f}}^a,
    K_{w}^a=h_w W_{K_{w}}^a, 
    V_{w}^a=h_w W_{V_{w}}^a,
\end{equation}
where $ a \in \{t, s\}$. $W_{Q_{w}}^{t}$ projects traffic embedding, $W_{K_{w}}^{t}$ and $W_{V_{w}}^{t}$ project weather embedding.  The temporal and spatial cross-attention scores can be formulated as:
\begin{equation}
A_{w}^{a } = \text{Softmax} \left(\frac{Q_{w}^{ a } K_{w}^{a\top}}{\sqrt{d_{h_f}}}\right).
\end{equation} 
Then, we get the weather-weighted flow embedding ${h}_f^{ t } = A_{w}^{t} V_{w}^{t} $ and ${h}_f^{ s } = A_{w}^{s} V_{w}^{s} $. 

We denote above weather-aware cross-attention block as:
\begin{equation} 
   {h}_f^{\text{weat}} = \text{CrossAttn} ( {h}_f, {h}_w).
\end{equation}

\subsubsection{Dual ST Memory Augmentation.}
Urban traffic patterns exhibit strong recurrence across time and space, such as rush hours or holiday-induced flow transitions. Moreover, even weather-induced anomalies often follow latent regularities. To exploit these recurring patterns, we introduce a learnable spatio-temporal memory bank that explicitly stores representative intrinsic and weather-influenced traffic states, enabling the model to retrieve relevant historical knowledge during prediction. 
Each branch is equipped with its own memory bank, where $M^{\text{intr}} \in \mathbb{R}^{L_{m} \times d_{m}}$ stores typical intrinsic spatio-temporal patterns, and $M^{\text{weat}} \in \mathbb{R}^{L_{m} \times d_{m}}$ captures traffic responses under varying weather conditions.
We linearly project $h_f^{\text{intr}}$ and $h_f^{\text{weat}}$ to obtain the query vector:
\begin{equation}  
    {q}^{{c}} =Q^{{c}}_m h_f^{{c}}+b^{{c}}, \ 
\end{equation}
where $c \in \{ \text{intr},\text{weat}\}$, $Q^{{c}}_m$ and $b^c$ are trainable parameters. 
Then ${q}^{{c}}$ acts as a
query to retrieve and fuse the most similar patterns from the corresponding memory: 
\begin{equation}  
{h}^{{c}} =A^{c } M^c, \ A^{c } = \text{Softmax} \left(q^c  M^{\text{c}\top} \right).
\end{equation}
This mechanism facilitates knowledge reuse, particularly when encountering rare events, thereby improving the model’s robustness.

\subsubsection{Weather Discrimination.} 
While \textit{W-STEnc} is designed to capture weather-aware traffic dynamics, \textit{I-STEnc} focuses on modeling traffic patterns that are invariant to weather conditions. To enforce this separation, the \textit{I-STEnc} output $h_f^{\text{intr}}$ is fed into a domain discriminator with a Gradient Reversal Layer (GRL)~\cite{ganin2015unsupervised}, encouraging weather-invariant representations. The discriminator is trained to predict the weather condition label ${c}_w$, while the GRL inverts gradients during backpropagation, encouraging \textit{I-STEnc} to suppress weather-discriminative information. Specifically, the discriminator outputs  $\hat{c}_w$ and we compute cross-entropy loss: 
\begin{gather}
    \hat{c}_w = \psi_{\text{w}}(\text{GRL}(h_f^{\text{intr}})), \\
    \mathcal{L}_{\text{dis}} = \text{CE}(\hat{c}_w, c_w).
\end{gather}
This adversarial setup enforces weather-invariant representations in \textit{I-STEnc}, strengthening its disentanglement from \textit{W-STEnc}.

\subsubsection{Adaptive Fusion and Prediction.}
The relative influence of intrinsic traffic patterns and weather-induced disturbances varies across contexts. In regular conditions, intrinsic patterns dominate, while in adverse weather, external factors may become decisive. Rigidly combining both branches equally may lead to suboptimal predictions. Hence, we introduce a dynamic fusion mechanism that adaptively allocates importance between the two branches. 
We design a gating module that receives the outputs of both branches and computes an adaptive weight vector $\alpha$ via a sigmoid activation, representing the relevance of weather-induced information. The fused representation is a weighted sum and then fed into an MLP predictor to produce the traffic forecast $\hat{y}$:
\begin{gather} 
{h^\text{fuse}} = \alpha \cdot h_f^{\text{intr}} + (1 - \alpha) \cdot h_f^{\text{weat}},\\ 
\hat{y} = \text{MLP}( {h^\text{fuse}} ). 
\end{gather} 
This adaptive fusion enables responsiveness to diverse traffic patterns and external perturbations. Given ground truth $y$, we optimize the model using MAE loss:
\begin{equation}
    \mathcal{L}_{\text{pre}} = \text{MAE}(\hat{y}, y).
\end{equation}
The overall optimization objective is:
\begin{equation}
    \mathcal{L}  = \mathcal{L}_{\text{pre}} +  \eta\mathcal{L}_{\text{dis}},
\end{equation}
where $\eta$ weights the adversarial weather classification loss.

\subsection{Spatio-temporal Causal Augmentation}
\subsubsection{Causal Support.}
Data scarcity during extreme weather events poses a major challenge for urban traffic flow modeling, often leading models to exploit spurious spatio-temporal correlations during training. For instance, models trained mainly on normal weather data may learn strong dependencies between residential and business districts during rush hours. However, such patterns do not hold under heavy rainfall, when remote work reduces commuting intensity or shifts peak periods. Similarly, while weekend mobility between residential areas and outdoor parks may appear correlated, rainy conditions often redirect movement toward indoor venues, breaking these learned associations.
This scarcity exposes a core limitation: models tend to learn distribution-specific statistical patterns rather than true causal relationships. When the distribution shifts, these statistical patterns collapse, and models inadvertently rely on coincidental co-occurrences that impede generalization under unseen or rare conditions.

\begin{figure}[!ht]
	\centering
   {\includegraphics[width=.99\columnwidth]{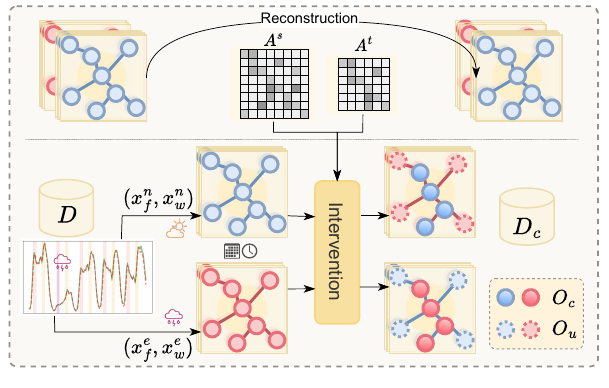}} 
 \caption{Spatio-temporal causal augmentation.}
 \label{fig:causal-aug}
\end{figure}

Recent studies have attempted to address this issue by identifying stable causal relations and perturbing irrelevant parts of the input to enforce the \textit{Invariance of Causality}~\cite{buhlmann2020invariance}. Following this paradigm, we perform causal discovery and controlled augmentation on spatio-temporal data, perturbing non-causal components across both spatial and temporal dimensions.  
Such perturbations break environment-specific spurious correlations while preserving causal mechanisms and improve generalization in atypical scenarios.

Formally, we examine historical observations $\mathcal{X}$ (traffic ${X}$ and weather ${M}$), future ground-truth traffic $\mathcal{Y}$, causal parts $\mathcal{X}_c$, and non-causal parts $\mathcal{X}_{nc}$ within a Structural Causal Model (SCM):
\begin{itemize}
    \item $\mathcal{X}_c \leftarrow \mathcal{X} \rightarrow \mathcal{X}_{nc}$;
    \item $\mathcal{X}_c \rightarrow \mathcal{Y} \dashleftarrow \mathcal{X}_{nc}$, where only $\mathcal{X}_c$ exerts true causal influence on $\mathcal{Y}$, while $\mathcal{X}_{nc}$ introduces spurious associations that may mislead prediction.
\end{itemize}
To eliminate confounding induced by $\mathcal{X}_{nc}$, we apply \textit{do-calculus} on $\mathcal{X}_c$ to block backdoor path $\mathcal{X}_{nc} \dashrightarrow \mathcal{Y}$: 
\begin{equation}
    P(\mathcal{Y}\mid do(\mathcal{X}_c))=\sum_{i=1}^{\tau} P(\mathcal{Y}\mid \mathcal{X},\mathcal{X}_{nc}) P(\mathcal{X}_{nc}=\mathcal{X}^i_{nc}).
\end{equation}
This formulation formalizes the causal intervention on $\mathcal{X}_c$, ensuring that the influence of non-causal variables is marginalized out. In practice, this insight enables us to perturb $\mathcal{X}_{nc}$ while keeping $\mathcal{X}_c$ intact in order to enforce invariance across environments. The model is therefore trained to rely on the stable causal components and to disregard environment-specific artifacts, ultimately learning spatio-temporal patterns that remain robust under extreme or previously unseen conditions.

\subsubsection{Causal Identification.}
\label{sec:causal_identification} 
To facilitate causal discovery, we use attention mechanisms to identify causal and non-causal parcels and time steps as shown in Fig.~\ref{fig:causal-aug}. Specifically, we perform a pre-experimental reconstruction using WED-Net and extract spatial and temporal attention maps to locate influential dependencies for each parcel and timestep. Unlike previous methods that assume spatially global causal regions, we argue that each parcel has its own context-specific causal and irrelevant regions. For example, weekday traffic in business areas may correlate with residential zones, while flows near train stations align with hotel areas.

Given a sample's traffic and weather data $(x_f^e, x_w^e)$, we obtain spatial self-attention maps from traffic ($A^s_f$) and cross-attention maps from weather to traffic ($A^s_w$). For each target parcel $v_i$, the $i$-th column of $A^s$ reflects the influence of all other parcels’ traffic or weather on the traffic of $v_i$. We select the top $r_{A^s}$ proportion as causal neighbors:
\begin{equation}
    O^{s}_{c}(v_i) = O^{s}_{c_f}(v_i) \cup O^{s}_{c_w}(v_i),
\end{equation}
and define the remaining as non-causal $O^s_u(v_i)$.

Similarly, we use temporal attention maps $A^t_f$ and $A^t_w$ to identify influential time steps for each $t_i$:
\begin{equation}
O^{t}_{c}(t_i) = O^{t}_{c_f}(t_i) \cup O^{t}_{c_w}(t_i).
\end{equation}
To preserve short-term continuity, we expand $O^t_c(t_i)$ with a fixed-size temporal window around $t_i$. All remaining time steps constitute the non-causal set $O^t_u(t_i)$.

\subsubsection{Causal Intervention.}
After identifying causal spatial parcels and temporal steps via attention mechanisms, we perform causal augmentation by perturbing non-causal parcels and time steps. Given an extreme-condition sample $(x_f^e, x_w^e)$ and a reference $(x_f^n, x_w^n)$, we replace the traffic and weather values of $O^s_u$ and $O^t_u$ in $(x_f^e, x_w^e)$  with those from $(x_f^n, x_w^n)$. Moreover, to ensure that the substitution of non-causal parts does not disrupt the original periodic temporal patterns, $(x_f^n, x_w^n)$ is chosen to match the day type (weekday or weekend) and the hour index of $(x_f^e, x_w^e)$.
We generate $r$ augmented samples for each extreme-condition sample, thereby transforming the original training set $D$ into the augmented set $D_c$. This preserves causal structure while varying irrelevant context, guiding the model to focus on true causal relationships.

\section{Experiments}
\subsection{Experimental Settings}
\subsubsection{Datasets.}
We collect representative human mobility data of New York City\footnote{{www.nyc.gov/site/tlc/about/tlc-trip-record-data.page}},  Chicago\footnote{{data.cityofchicago.org}}, and Washington DC\footnote{{opendata.dc.gov/search?q=taxi}} to evaluate the effectiveness of our model. The taxi data includes the timestamps and location parcel IDs of pick-up and drop-off for each trip. Meteorological data\footnote{{www.ncei.noaa.gov/maps/hourly}} is sourced from weather stations in each city, including hourly precipitation, wind speed, etc. Each land parcel’s meteorological data is estimated through inverse-distance weighted interpolation from nearby weather stations. We align the temporal and spatial granularity of traffic and weather data for each city. 
Table~\ref{data-statistic} presents detailed dataset statistics. The data is split into 50\%/25\%/25\% based on chronological order for training, validation, and testing, respectively. 
We prioritize precipitation to define the scope of weather conditions, since rainfall has a more direct and immediate impact on road conditions and travel demand than temperature or wind. Samples with average precipitation above 0.1 inches per hour are labeled extreme rainstorm cases, while those below are treated as normal weather samples; this threshold is chosen based on the standard definition of rainfall intensity\footnote{{en.wikipedia.org/wiki/Rain}} together with the empirical rainfall distributions of the three cities. Under this criterion, the training and test sets cover both normal and extreme weather conditions, but extreme-condition samples remain relatively scarce. We partition the test set into normal and extreme conditions for separate evaluation.

\begin{table}[!h]  \small
    \centering 
    \caption{Dataset statistics.}
    \label{data-statistic}
     \resizebox{\linewidth}{!}{
        \begin{tabular} { ccc cc c c}
            \toprule  
            {City} &  \makecell[c]{\#Parcels}&  \makecell[c]{Time \\Span}  &\makecell[c]{Weather \\ Condition}& \makecell[c]{\#Trian} & \makecell[c]{\#Valid} & \makecell[c]{\#Test}  \\
            \midrule
              \multirow{2}{*}{NYC}& \multirow{2}{*}{66}& \multirow{2}{*}{\makecell[c]{03/2017-\\10/2017}} &Normal&\multirow{2}{*}{3505}&1044&1056\\ 
                                  &                                 &                                      &Extreme&&365&290\\
            \midrule
              \multirow{2}{*}{CHI}& \multirow{2}{*}{77}& \multirow{2}{*}{\makecell[c]{03/2017-\\10/2017}} &Normal &\multirow{2}{*}{3505}&1085&997\\  
          
                                  &                                 &                                       &Extreme&&282&329\\
            \midrule
              \multirow{2}{*}{DC}& \multirow{2}{*}{69}& \multirow{2}{*}{\makecell[c]{01/2017-\\08/2017}} &Normal&\multirow{2}{*}{3476}&1077&1038\\ 
          
                                  &                                 &                                      &Extreme&&353&384\\
            \bottomrule
        \end{tabular}
    }
    
\end{table}

\begin{table*}[ht!]
\centering
\caption{Model performance comparison under extreme and normal weather across NYC, CHI, and DC.}
\label{tab:overall-performance}
\begin{tabular}{lcccccccccccc}
\toprule
\multirow{3}{*}{\textbf{Method}}&  \multicolumn{4}{c}{\textbf{NYC}} 
& \multicolumn{4}{c}{\textbf{CHI}} 
& \multicolumn{4}{c}{\textbf{DC}} \\
\cmidrule(lr){2-5} \cmidrule(lr){6-9} \cmidrule(lr){10-13}
 & \multicolumn{2}{c}{Extreme} & \multicolumn{2}{c}{Normal}
  & \multicolumn{2}{c}{Extreme} & \multicolumn{2}{c}{Normal}
  & \multicolumn{2}{c}{Extreme} & \multicolumn{2}{c}{Normal} \\
\cmidrule(lr){2-3} \cmidrule(lr){4-5}
\cmidrule(lr){6-7} \cmidrule(lr){8-9}
\cmidrule(lr){10-11} \cmidrule(lr){12-13}
&  MAE & RMSE & MAE & RMSE 
  & MAE & RMSE & MAE & RMSE 
  & MAE & RMSE & MAE & RMSE \\
   \hline
   \multicolumn{13}{c}{{No Augmentation}}\\ 
  \hline
DCRNN & 0.7221 & 1.0465 & 0.7351 & 1.0549 & 0.1562 & 0.7286 & 0.1591 & 0.7284 & 0.2450 & 0.7843 & 0.2515 & 0.8003 \\
AGCRN & 0.6732 & 0.9900 & 0.6768 & 0.9910 & 0.2118 & 1.0024 & 0.2192 & 1.0115 & 0.2764 & 0.9287 & 0.2836 & 0.9572 \\
GTS & 0.5317 & 0.7886 & 0.5319 & 0.7852 & 0.2252 & 0.9460 & 0.2277 & 0.9507 & 0.2798 & 0.8873 & 0.2830 & 0.9119 \\
STGCN & 0.3626 & 0.6121 & 0.3530 & 0.5944 & 0.1059 & 0.5183 & 0.1028 & 0.4970 & 0.1701 & 0.5773 & 0.1655 & 0.5511\\
GWNet & 0.1762 & 0.3164 & 0.1716 & 0.3051 & 0.0544 & 0.2207 & 0.0535 & 0.2125 & 0.1015 & 0.2933 & 0.0970 & 0.2738 \\
STAEformer & 0.1611 & 0.2817 & 0.1503 & 0.2619 &0.0532	&0.1904	&0.0390	&0.1901 & 0.0976 & 0.2983 & 0.0940 & 0.2789 \\
MTGNN & 0.1344 & 0.2494 & 0.1304 & 0.2394 & 0.0398 & 0.1639 & 0.0370 & 0.1496 & 0.0881 & 0.2598 & \textbf{0.0848} & \textbf{0.2392} \\
WED-Net & \textbf{0.1281} & \textbf{0.2330} & \textbf{0.1206} & \textbf{0.2121} & \textbf{0.0386} & \textbf{0.1558} & \textbf{0.0358} & \textbf{0.1468} & \textbf{0.0871} & \textbf{0.2583} & 0.0887 & 0.2556 \\

\hline
 \multicolumn{13}{c}{{Causal Augmentation}} \\
 \hline
DCRNN & 0.4388 & 0.7385 & 0.4357 & 0.7279 & 0.1131 & 0.5353 & 0.1144 & 0.5292 & 0.2059 & 0.7320 & 0.2094 & 0.7485 \\
AGCRN & 0.6701 & 0.9893 & 0.6766 & 0.9901 & 0.2117 & 1.0019 & 0.2192 & 1.0110 & 0.2761 & 0.9276 & 0.2833 & 0.9561 \\
GTS & 0.4999 & 0.7396 & 0.4964 & 0.7267 & 0.2029 & 0.8600 & 0.2062 & 0.8661 & 0.2703 & 0.7335 & 0.2797 & 0.7751 \\
STGCN & 0.3614 & 0.6123 & 0.3527 & 0.5977 & 0.0980 & 0.4760 & 0.0943 & 0.4559 & 0.1636 & 0.5464 & 0.1605 & 0.5277\\
GWNet & 0.1750 & 0.3098 & 0.1688 & 0.2965 & 0.0501 & 0.2123 & 0.0489 & 0.2021 & 0.0998 & 0.2902 & 0.0917 & 0.2704  \\
STAEformer & 0.1568 & 0.2861 & 0.1408 & 0.2522 & 0.0431 & 0.1680 & 0.0400 & 0.1624 & 0.0950 & 0.2929 & 0.0910 & 0.2701 \\
MTGNN & 0.1341 & 0.2465 & 0.1257 & 0.2241 & 0.0380 & \textbf{0.1522} & 0.0353 & \textbf{0.1439} & 0.0920 & 0.2841 & 0.0862 & 0.2540 \\
WED-Net & \textbf{0.1172} & \textbf{0.2087} & \textbf{0.1152} & \textbf{0.2059} & \textbf{0.0372} &  0.1541 & \textbf{0.0347} & 0.1459 & \textbf{0.0863} & \textbf{0.2506} & \textbf{0.0825} & \textbf{0.2297} \\

\bottomrule
\end{tabular}

\end{table*}

\subsubsection{Baselines.}
To evaluate the performance of WED-Net, we compare it with widely used baselines in urban spatio-temporal traffic prediction, including spatiotemporal graph neural networks (STGCN~\cite{10.5555/3304222.3304273}, DCRNN~\cite{li2018diffusion}, AGCRN~\cite{bai2020adaptive}, GWNet~\cite{10.5555/3367243.3367303}, GTS~\cite{shang2021discrete}, MTGNN~\cite{10.1145/3394486.3403118}), and Transformer-based models 
(STAEformer~\cite{10.1145/3583780.3615160}).
To further validate the effectiveness of our proposed causal augmentation approach, we integrate it into above baselines and evaluate the performance gains.

\subsubsection{Hyperparameter and Reproducibility.}
For WED-Net’s embedding layer, $E_h$ uses 12 hidden features, while $E_{a_s}$ and $E_{a_t}$ use 18 each. The periodic time-of-day and day-of-week embeddings $E_p$ use 12 features each. Both \textit{I-STEnc} and \textit{W-STEnc} contain 4 Transformer blocks with 4-head multi-head attention.
To ensure fairness, all models use both traffic and weather inputs. 
We use batch size 128 and learning rate 1e-3 with a OneCycleLR scheduler and AdamW optimizer with weight decay 5e-4. Experiments are conducted on NVIDIA A40 GPUs with Ubuntu 20.04.6 LTS, using Python 3.9.12, PyTorch 1.13.0, and DGL 1.1.3.

\subsection{Performance Evaluation}

\subsubsection{Overall Evaluation.}
We focus on urban flow prediction in different weather conditions. During testing, we assess the model's performance under both extreme and normal weather conditions for each city.  We use historical taxi flow and weather from the past 12 time steps to predict the flow for the next 12 time steps.

As shown in Table~\ref{tab:overall-performance}, WED-Net achieves the lowest MAE and RMSE across most cities and weather conditions, indicating its superior generalization and robustness.  In particular, the performance advantage is most pronounced in extreme weather, where traffic dynamics exhibit greater irregularity.
The performance advantage is most pronounced under extreme weather, with improvements of 14\% and 9\% over the second-best model MTGNN in NYC and DC, respectively.
Compared with baselines that also leverage spatio-temporal dependencies, WED-Net benefits from three architectural innovations. First, the dual-branch disentanglement module enables independent modeling of weather-sensitive and weather-invariant patterns, which avoids mutual interference and supports specialized learning. Second, the memory bank augmentation modules provide representative historical knowledge for both intrinsic and weather-induced branches, improving the model’s ability to cope with rare or unseen events. Third, the adaptive fusion mechanism allows the model to dynamically adjust the influence of each branch, resulting in better responsiveness under different conditions.

In addition to architectural design, the causal augmentation strategy further enhances generalization. By perturbing the non-causal parts of flow and weather signals while preserving causally relevant ones, the model is encouraged to focus on core causal factors. 
As shown in the lower half of Table~\ref{tab:overall-performance}, causal augmentation benefits most models, with DCRNN improving by 25\%, GTS by 8\%, and STGCN, GWNNet, STAEformer, and WED-Net by around 5\%, while WED-Net consistently maintains the best performance.
This suggests our disentangled design captures stable dependencies, with causal augmentation further enhancing generalization.

\begin{figure}[!ht]
	\centering 

    \subfloat[\label{fig:causal_parcel_vis_intr}Intrinsic causal neighbors.] 
     {\includegraphics[width=.9\columnwidth]{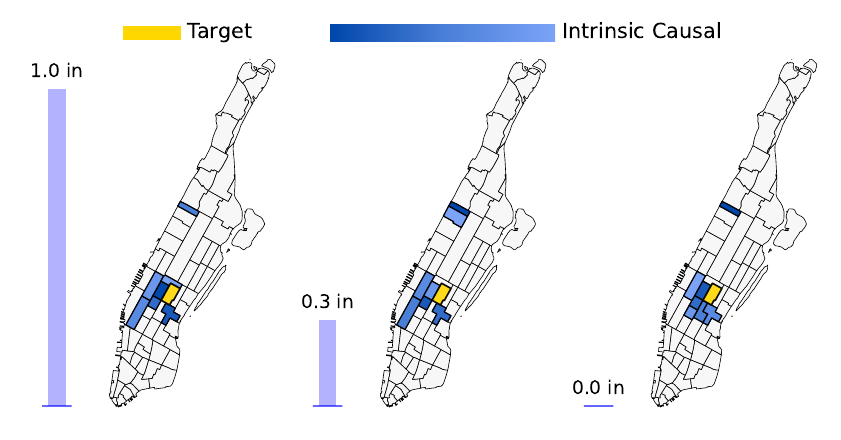}}  \\
      \subfloat[\label{fig:causal_parcel_vis_weat}Weather-effect causal neighbors.]
     {\includegraphics[width=.9\columnwidth]{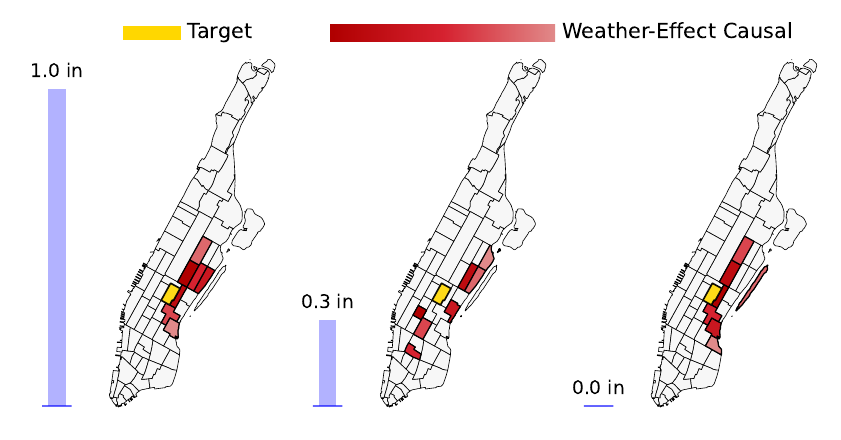}} 
\caption{Intrinsic and Weather-effect causal neighbors of parcel $v_{37}$ under different weather conditions.}
 \label{fig:causal_parcel_vis}
\end{figure}

Overall, these results show that WED-Net effectively disentangles and fuses heterogeneous dependencies in urban flow prediction, remaining robust under normal and extreme weather.

\subsubsection{Intrinsic and Weather-Effect Causal Neighbors Visualization.}
Fig.~\ref{fig:causal_parcel_vis} further illustrates the effectiveness of weather-effect disentanglement and causal identification method. We randomly select a target parcel $v_{37}$ (yellow) and visualize three samples at 10:00 on a Friday, each from a different date, under precipitation levels of 1.0, 0.3, and 0.0 inches per hour, respectively. Using the causal identification procedure described in Sec.~\ref{sec:causal_identification}, we plot the intrinsic causal neighbors $O^{s}_{c_f}(v_{37})$ (blue) in Fig.~\ref{fig:causal_parcel_vis_intr}, and the weather-effect causal neighbors $O^{s}_{c_w}(v_{37})$ (red) in Fig.~\ref{fig:causal_parcel_vis_weat}. The color intensity indicates the relative causal importance.
As shown, intrinsic causal neighbors stay largely consistent across weather conditions, while weather-effect causal neighbors vary markedly with precipitation, indicating that \textit{I-STEnc} and \textit{W-STEnc} disentangle intrinsic dependencies from weather-induced modulations and that the causal identification is stable and accurate.

\begin{figure}[!ht]
	\centering
 \subfloat[\label{draw-pred}]{\includegraphics[width=.666\columnwidth]{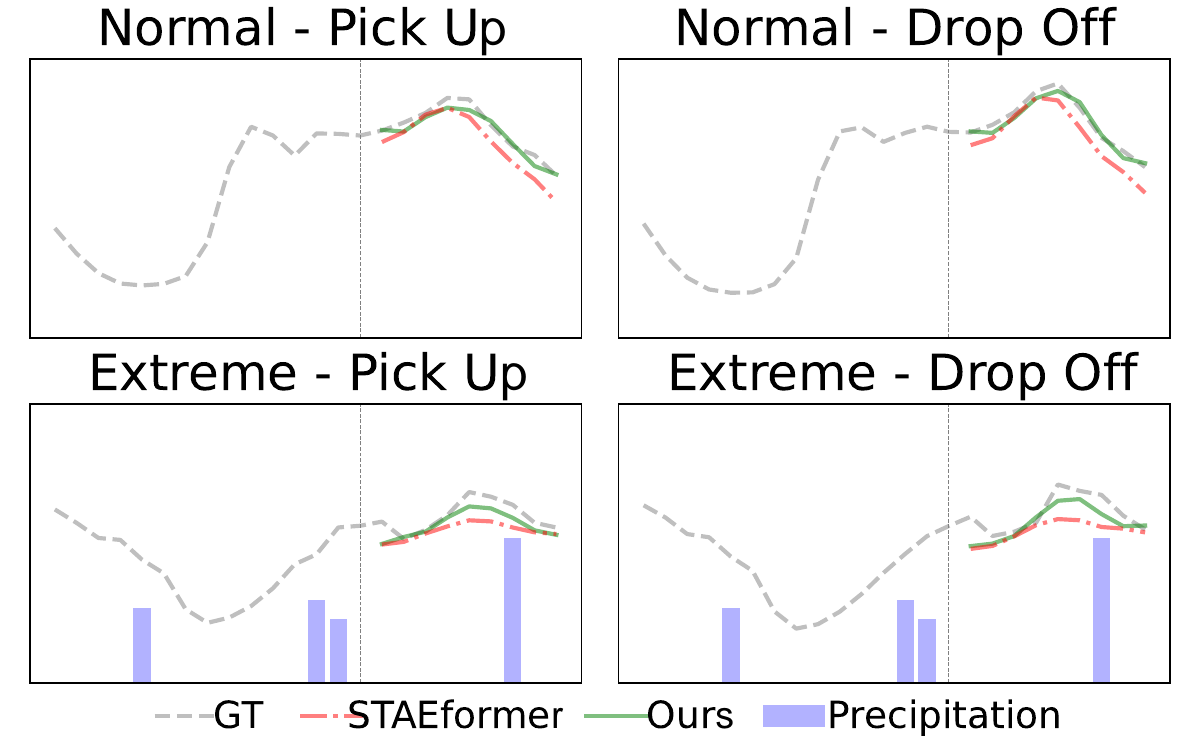}} 
  \subfloat[\label{PCA}]{\includegraphics[width=.333\columnwidth]{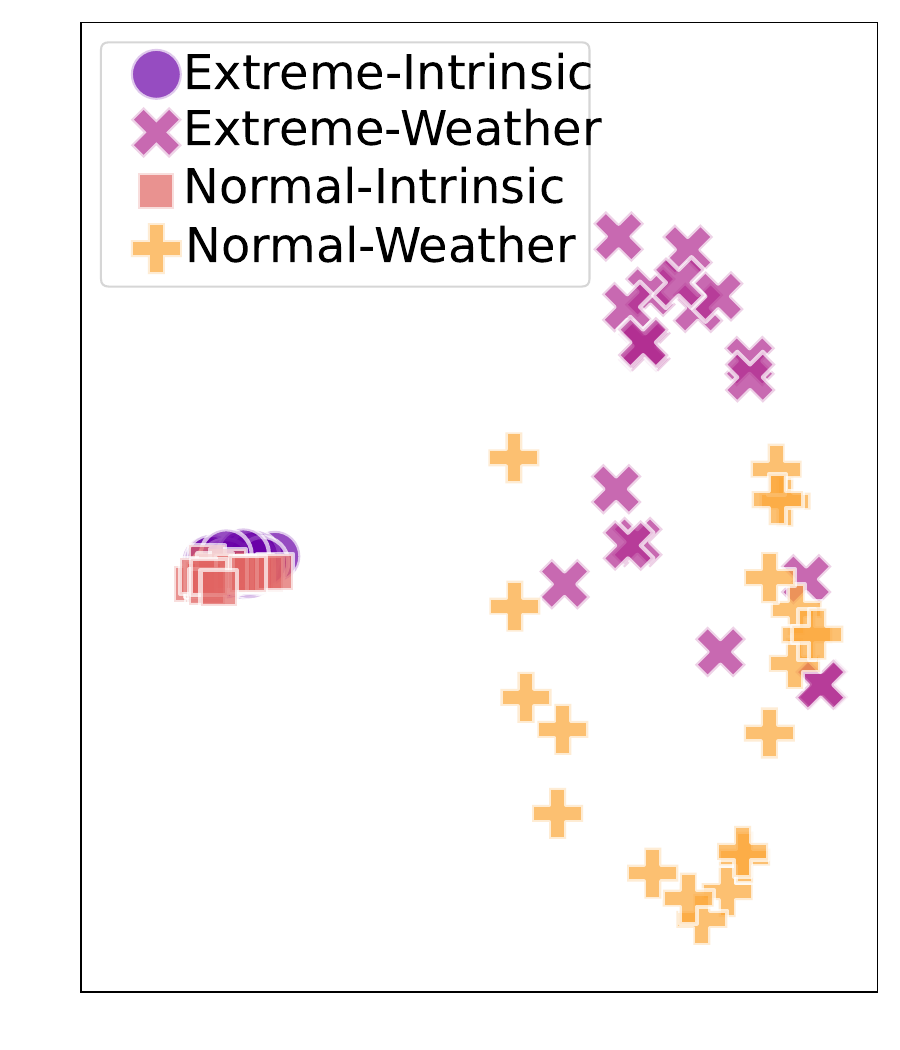}} 
 \caption{(a) Taxi flow prediction under varying weather; (b) Latent representation visualization of our model across weather scenarios.}
\end{figure}

\subsubsection{Weather-Aware Flow Pattern Visualization.}
To evaluate WED-Net’s prediction performance under different weather conditions, Fig.~\ref{draw-pred} compares its prediction with those of STAEformer. Under normal weather (top row), the two models perform similarly. Under heavy rain (bottom row), STAEformer exhibits clear deviations during intense precipitation, whereas WED-Net better tracks the true flow trend, demonstrating greater robustness to extreme conditions.
Fig.~\ref{PCA} further visualizes the hidden representations learned by WED-Net. We apply PCA to the hidden features from both \textit{I-STEnc} and \textit{W-STEnc} for samples under extreme and normal conditions. Intrinsic features form clusters that are largely invariant to weather, indicating that the model successfully learns weather-invariant intrinsic traffic patterns. In contrast, weather-induced features exhibit clear separation across different weather conditions, confirming that WED-Net effectively disentangles weather-driven dynamics from intrinsic traffic patterns.

\subsubsection{Ablation Study.} 
To evaluate the effectiveness of each module in our framework, we conduct comprehensive ablation studies on both model components and the proposed causal augmentation strategy. 
We design several ablation variants to assess key components of the model. The weather input and its corresponding branch is entirely removed (Ours-w), or replaced with self-attention to disable weather-flow interaction while retaining weather input (Ours-wca). ST memory augmentation method is removed (Ours-mem), and weather discrimination loss is dropped (Ours-ddl). 
For causal augmentation, we degrade spatial causal parcel localization by using adjacent nodes from the adjacency matrix (CaAu-s), and degrade the selection of temporal causal steps to using all time steps (CaAu-t).

\begin{figure}[!ht]
	\centering 
     {\includegraphics[width=.49\columnwidth]{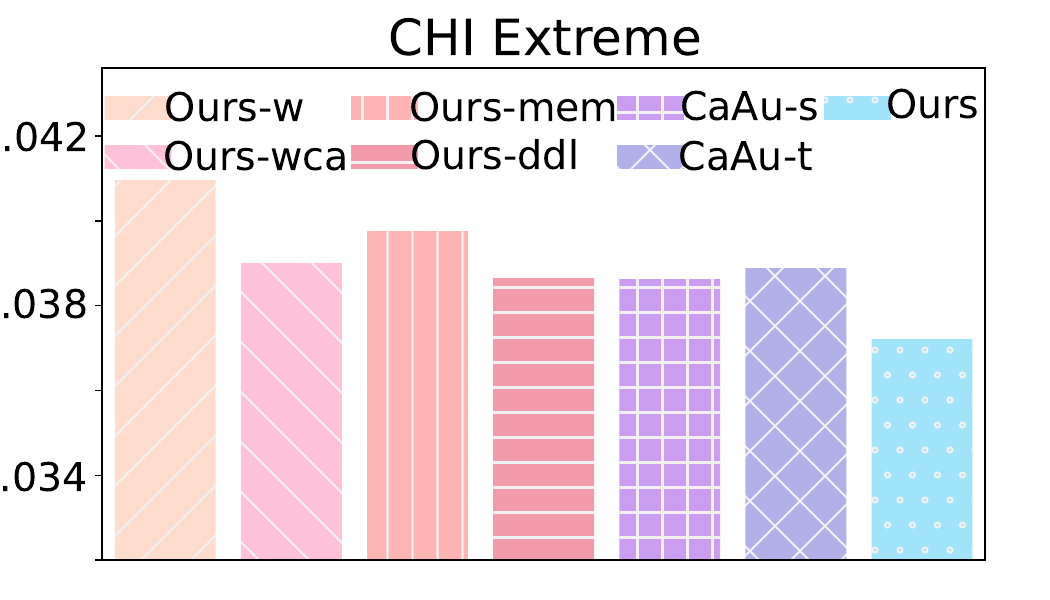}} 
     {\includegraphics[width=.49\columnwidth]{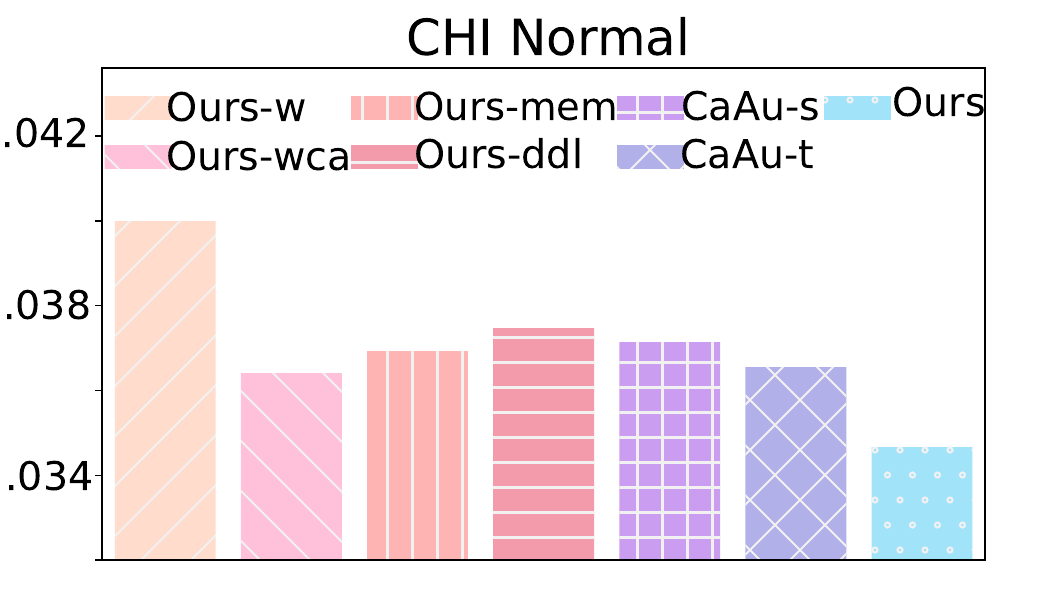}} \\ 
     {\includegraphics[width=.49\columnwidth]{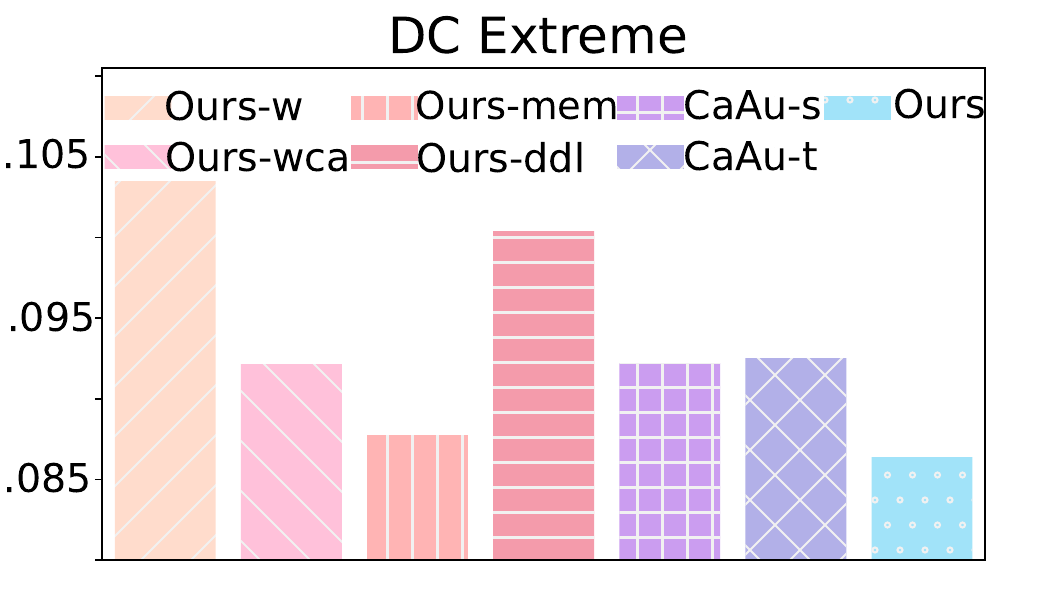}} 
     {\includegraphics[width=.49\columnwidth]{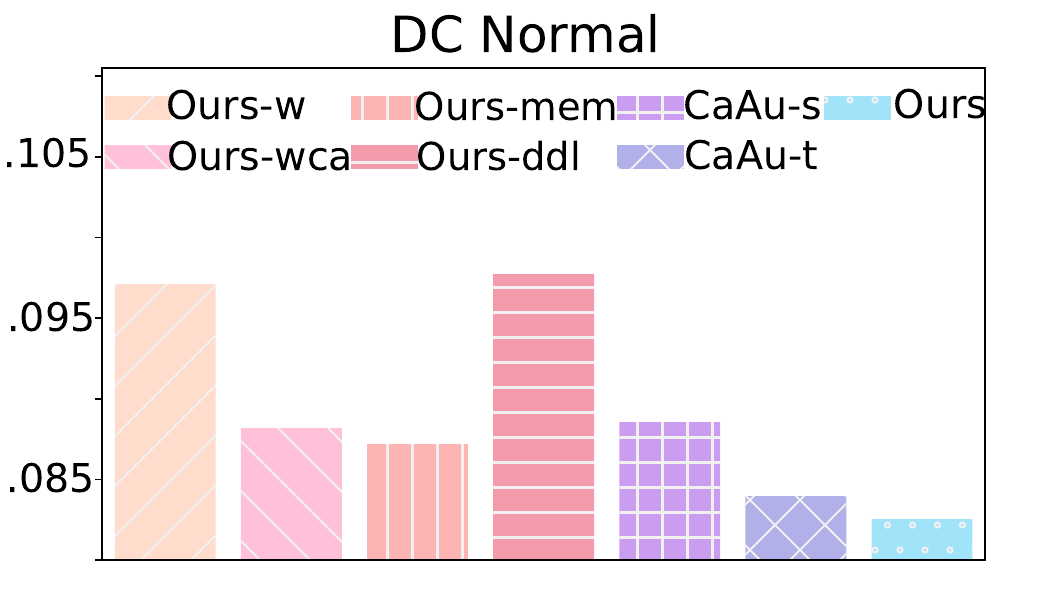}} \\ 

\caption{Ablation results on CHI and DC under extreme and
normal weather conditions.}
 \label{fig:ablation}
\end{figure}

As shown in Fig.~\ref{fig:ablation}, removing the whole weather modality leads to the largest performance drop, especially under extreme weather, underscoring the need for weather modeling. Simply replacing cross-attention with self-attention also degrades performance, indicating that direct weather-flow interaction is crucial. Removing the memory module or the weather discrimination mechanism consistently harms performance, with the latter causing especially large drops on DC, highlighting its role in handling weather-sensitive scenarios. For causal augmentation, degrading any single type of causal localization leads to performance degradation, highlighting the effectiveness and complementary nature of different causal region discovery mechanisms.
 
\begin{figure}[!ht]
	\centering
   {\includegraphics[width=.325\columnwidth]{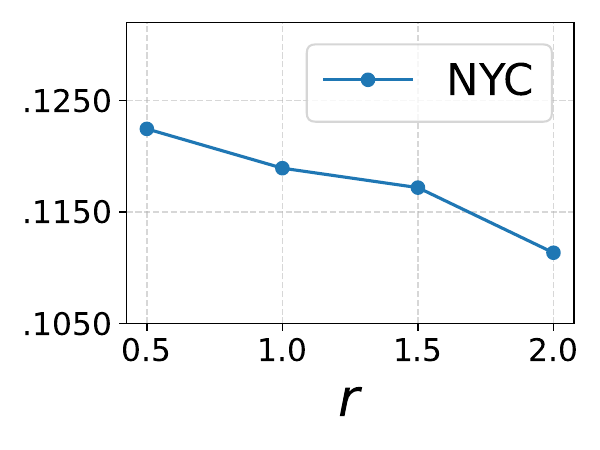}} 
     {\includegraphics[width=.325\columnwidth]{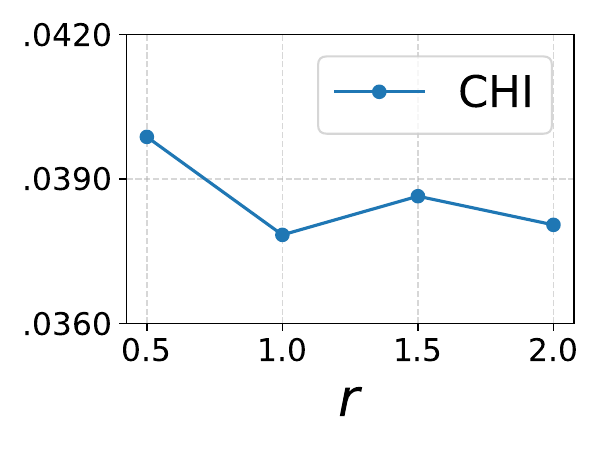}} 
   {\includegraphics[width=.325\columnwidth]{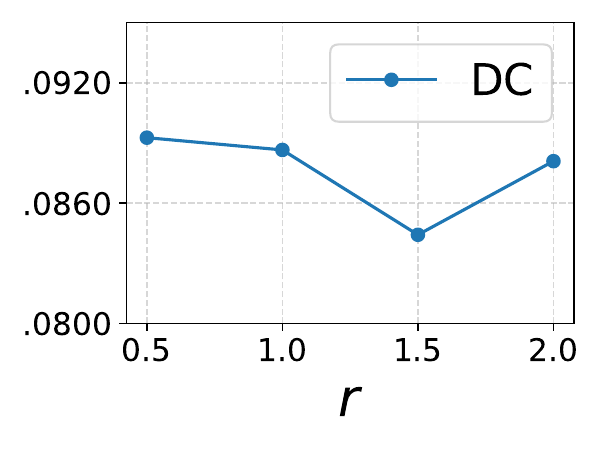}}  \\
     {\includegraphics[width=.325\columnwidth]{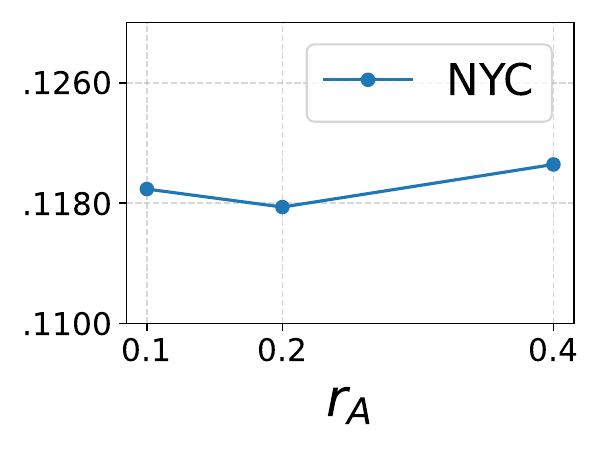}} 
   {\includegraphics[width=.325\columnwidth]{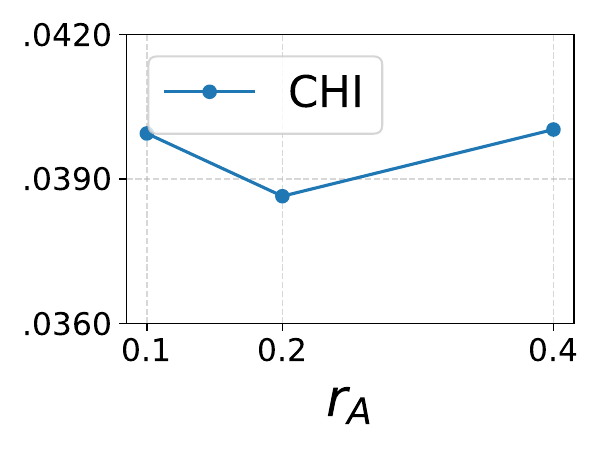}} 
    {\includegraphics[width=.325\columnwidth]{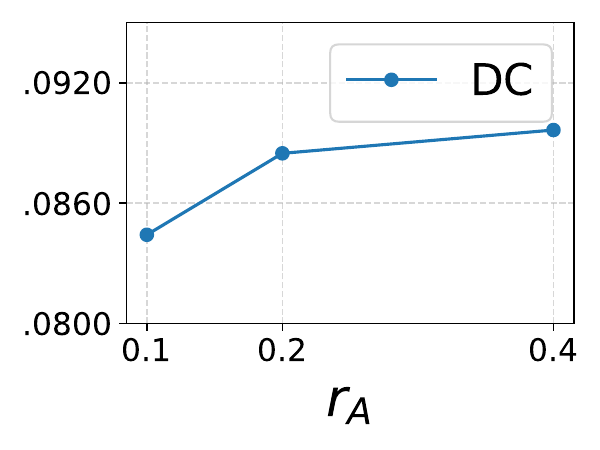}}  \\
     {\includegraphics[width=.325\columnwidth]{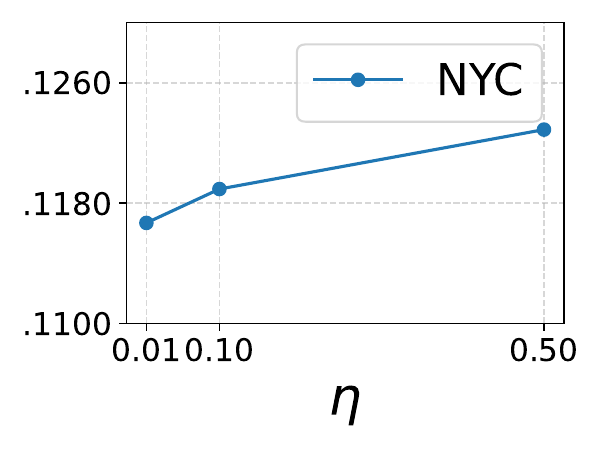}} 
   {\includegraphics[width=.325\columnwidth]{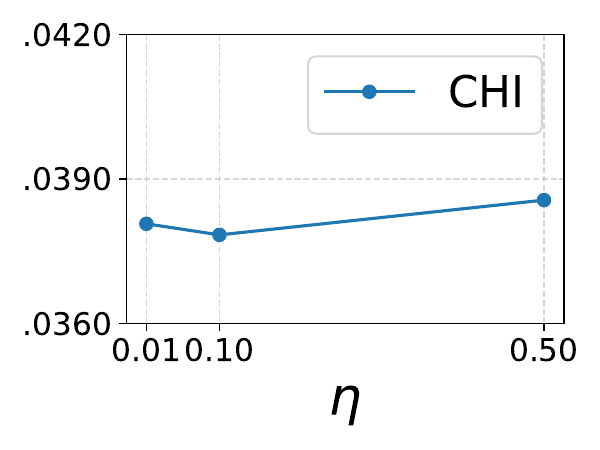}} 
    {\includegraphics[width=.325\columnwidth]{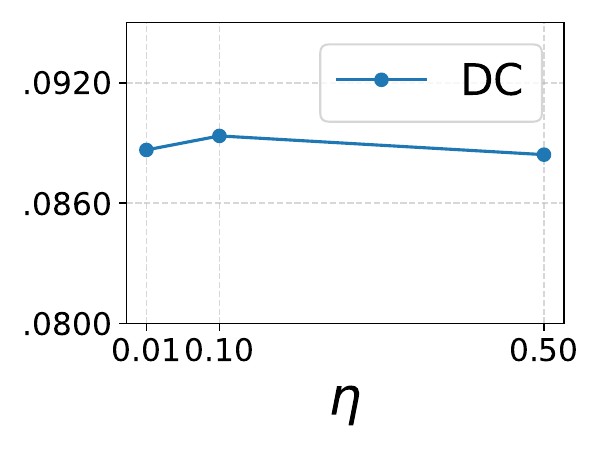}}  \\
 \caption{Parameter sensitivity analysis on $r$, $r_A$, and $\eta$.}
 \label{fig:sensi}
\end{figure}

\subsubsection{Parameters Sensitivity.} 
We further analyze the sensitivity of three key hyperparameters: $r$, $r_A$, and $\eta$, as shown in Fig.~\ref{fig:sensi}. Overall, the model maintains stable performance across a wide range of settings.
Increasing $r$ generally improves performance on NYC and DC, indicating the benefit of stronger data augmentation. For $r_A$, lower values (e.g., 0.1 or 0.2) tend to perform better, suggesting that replacing more non-causal regions contributes to better generalization. As for $\eta$, small values (e.g., 0.01 or 0.1) consistently yield better results, while a larger $\eta = 0.5$ slightly degrades performance on NYC and CHI, likely due to overemphasizing the auxiliary discrimination objective. These results validate the robustness of our method under different configurations.

\section{Conclusion}
We propose WED-Net, a dual-branch spatio-temporal forecasting model that explicitly disentangles intrinsic traffic patterns from weather-induced effects. By incorporating dedicated memory modules and a weather discriminator, WED-Net is able to capture both stable, environment-invariant dynamics and fine-grained, weather-sensitive variations in mobility. When combined with the proposed causal augmentation strategy, WED-Net generalizes effectively under distribution shifts and consistently outperforms competitive baselines, with particularly pronounced gains under extreme weather conditions. These capabilities make WED-Net a practical and reliable tool for robust urban mobility forecasting and for supporting data-driven disaster preparedness and response planning, thereby enhancing urban resilience.

\section*{Acknowledgements}
This work was supported by the Public Computing Cloud at Renmin University of China and the Fund for Building World-Class Universities (Disciplines) at Renmin University of China.

\bibliographystyle{ACM-Reference-Format}
\balance
\bibliography{sample-base}

\appendix

\end{document}